\documentclass[letterpaper, 10 pt, conference]{ieeeconf}  

\IEEEoverridecommandlockouts                              

\usepackage{epsfig} 
\usepackage{amsmath} 
\usepackage{amssymb}  

\makeatletter 
\let\NAT@parse\undefined
\makeatother
\usepackage{hyperref} 
\usepackage{dirtytalk} 
\DeclareMathOperator*{\argmin}{argmin} 
\DeclareMathOperator*{\GrD}{G \rightarrow D}

\title{\LARGE \bf
Relative Drone - Ground Vehicle Localization using LiDAR and Fisheye Cameras through Direct and Indirect Observations 
}

\author{Jan Hausberg$^{1}$, Ryoichi Ishikawa$^{2}$, Menandro Roxas$^{3}$, Takeshi Oishi$^{2}$
\thanks{$^{1}$Karlsruhe Institute of Technology, Karlsruhe, Germany
        {\tt\small uhelc@student.kit.edu}}%
\thanks{$^{2}$The University of Tokyo,
        Tokyo, Japan
        {\tt\small \{ishikawa, oishi\}@cvl.iis.u-tokyo.ac.jp}}%
\thanks{$^{3}$ Line Corporation, Tokyo, Japan
        {\tt\small menandro.roxas @linecorp.com}}%
}

\begin{document}

\maketitle
\thispagestyle{empty}
\pagestyle{empty}
\addtocounter{footnote}{+3}

\begin{abstract}
Estimating the pose of an unmanned aerial vehicle (UAV) or drone is a challenging task. 
It is useful for many applications such as navigation, surveillance, tracking objects on the ground, and 3D reconstruction. 
In this work, we present a LiDAR-camera-based relative pose estimation method between a drone and a ground vehicle, using a LiDAR sensor and a fisheye camera on the vehicle's roof and another fisheye camera mounted under the drone. 
The LiDAR sensor directly observes the drone and measures its position, and the two cameras estimate the relative orientation using indirect observation of the surrounding objects.
We propose a dynamically adaptive kernel-based method for drone detection and tracking using the LiDAR. 
We detect vanishing points in both cameras and find their correspondences to estimate the relative orientation. Additionally, we propose a rotation correction technique by relying on the observed motion of the drone through the LiDAR.
In our experiments, we were able to achieve very fast initial detection and real-time tracking of the drone.
Our method is fully automatic.
\end{abstract}

\section{Introduction}
Unmanned aerial vehicles, or drones, have numerous benefits when used in robotics applications such as surveillance, agriculture monitoring, and 3D reconstruction \cite{heritageitaly, heritagebridges}. 
The altitude of the drone allows for detecting multiple objects on the ground, which is difficult for ground-based robots due to natural occlusions and limited field-of-view (FOV). 
In a similar sense, GPS data from drones can be used for aiding during navigation and autonomous driving. 
Since drones can avoid tall buildings that can obstruct GPS signals, ground vehicles can rely on them for more accurate positioning. 

To benefit from the drone, we need an accurate estimation of the drone's pose.
Recent advances in Computer and Robot Vision allow us to detect and track drones even in complex scenes.
However, estimating the relative poses with an absolute scale is still a challenging task.
The vision-based approach has a limitation in estimating the relative pose between the two systems.
Due to the largely different viewpoints, targets in the scene can have different appearances or too small to be meaningful, especially for distant objects.
Moreover, direct observations solve only the baseline between the two systems, even if the sensors are visible to each other.

Therefore, we need to use indirect and common measurements, such as gravity direction,
landmarks such as the sun or shadow direction, etc. to solve the relative pose estimation.
However, the accuracy of gravity sensors may vary depending on the surrounding environment, and finding mutually visible landmarks from largely different viewpoints can be difficult.
Hence, we need to use more global visual features that can be easily observed in the scene.

In this paper, we present a method to address the relative pose problem through direct and indirect observations.
First, the drone's relative position, with respect to a ground vehicle, is solved using a LiDAR-based tracking system with a scanning mechanism.  
A kernel-based point-cloud processing method allows the system to detect and track the drone robustly and in real-time.
Second, the relative rotation between the drone and the ground vehicle is solved through the detection and robust alignment of the vanishing points (VPs) derived from their cameras' views. 
Our alignment method can recover the relative rotation in the middle of the deployment by reasoning on the drone and vehicle's relative motions.

\begin{figure}[t]
    \centering
    \includegraphics[width=.9\linewidth]{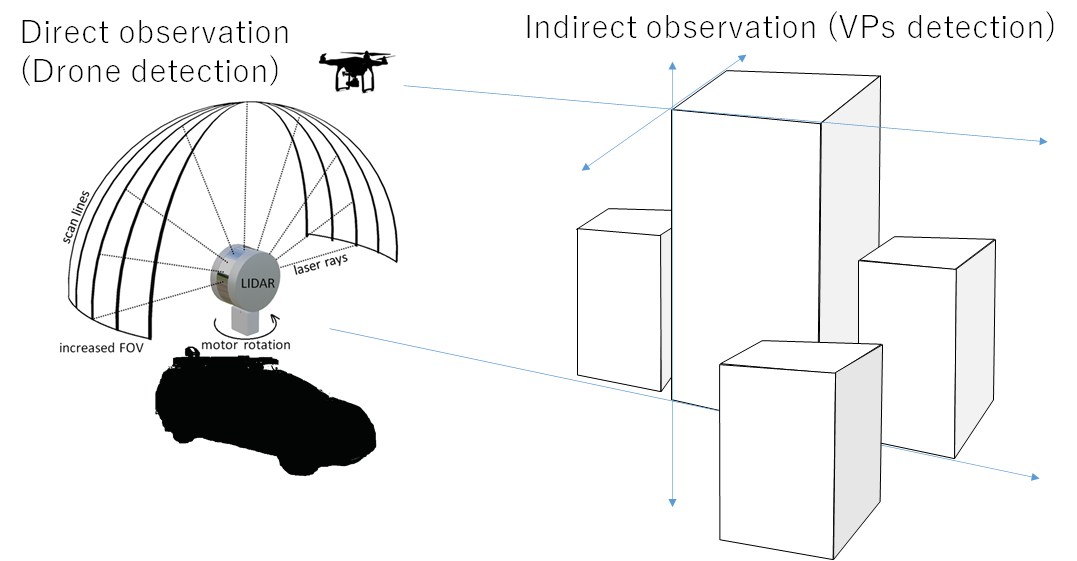}
    \caption{Overview of our proposed ground vehicle - relative drone pose estimation method. Using a LiDAR sensor on the vehicle's roof, we detect and track the drone's position (direct observation). Using two fisheye cameras (one on vehicle, one on drone), we detect and align VPs to estimate the drone's rotation (indirect observation).}
    \label{fig:overview}
\end{figure}

We summarize the contributions of the paper as follows:
\begin{itemize}
    \item We propose a relative pose estimation framework using a fusion of LiDAR and cameras with direct and indirect observations.
    \item We propose an adaptive kernel-based 3D detector for drone detection, followed by real-time drone tracking.
    \item We propose a robust relative orientation estimation method using VPs and relative motion. 
\end{itemize}
Our results show that our method can successfully detect and track the drone’s pose in real scenes.
\section{Related Work}
Several methods have been proposed for detecting drones using a multitude of modalities. Radar systems \cite{hommes} are complex and expensive. RF signal detection \cite{Boddhu2013}, which intends to capture the communication between the drone and the ground operator, may fail due to the surrounding environment. A low-cost version for drone detection can be achieved through visual imagery \cite{Rozantsev2017}\cite{Aker2017}\cite{Schumann2017}\cite{Unlu2019}, but the absolute range cannot be measured. Using LiDAR systems can overcome this problem. Hammer et al. \cite{Hammer2018} used four LiDAR sensors to detect and track a drone. In contrast, our system requires fewer resources by only using one LiDAR sensor and two fisheye cameras. Moreover, we solve both the position and orientation of the drone relative to the ground vehicle. 

Countless methods \cite{klein09cameraphone}\cite{caruso2015_omni_lsdslam}\cite{engel14eccv} have been proposed to solve the relative camera pose estimation. Most methods have relied on classical descriptors \cite{lowe_2004}\cite{bay_tuytelaars_van}\cite{murTRO2015} and local feature detection. These methods show inefficiency towards large viewpoint changes. Moreover, vision-only methods cannot give an absolute scale. In contrast, our method can handle largely different viewpoints and gives an absolute scale.

Using convolutional neural networks (CNNs), deep learning-based methods \cite{konda_memisevic_2015}\cite{deepvo}\cite{Melekhov2017} have been proposed to address the drawbacks of the hand-drawn descriptors while achieving better performance.
However, CNNs require large datasets and time-consuming training. In contrast, our system is fully unsupervised and, therefore, does not require training data.

For finding the relative orientation between distant cameras having significantly different views, vanishing points are more robust than feature detection. Caprile and Torre \cite{Caprile1990} proposed a method that uses VPs to calibrate a system consisting of multiple cameras and, thus, also finds the relative rotation between them by detecting corresponding VPs. Our rotation estimation approach is similar to \cite{Caprile1990} except that we can recover the orientation online by reasoning on the motion of the drone.
\section{Overview}
Given a ground vehicle (G) and a drone (D), our goal is to find the relative pose between G and D -- \textbf{g}round vehicle - \textbf{r}elative \textbf{d}rone (GrD) rotation $R_{G\rightarrow D} \in \mathbb{R}^{3 \times 3}$ and the GrD translation vector $t_{G\rightarrow D} \in \mathbb{R}^3$. 
We assume that the vehicle and the drone have calibrated cameras, which means that the variables pertaining to the drone and vehicle are in their camera coordinate systems. 
We also assume that the vehicle has a LiDAR sensor, which is also calibrated with the vehicle's camera.
An overview of our system is shown in Fig. \ref{fig:overview}.

Our method is organized as follows. In Sec. \ref{sec:drone_tracking}, we propose a GrD translation estimation method using drone detection and tracking with the LiDAR (direct observation). This results in a $t_{G\rightarrow D}$ estimate with absolute scale. 

In Sec. \ref{sec:vp_detection}, we present a robust $R_{\GrD}$ estimation by detecting and aligning the VPs between the cameras (indirect observation). In this case, we assume that the environment has dominant VPs, such that the cameras can detect the corresponding VPs even at largely different viewpoints. We ensure the accuracy of the VP correspondences by correlating the relative motion of the drone (self-relative) with the LiDAR detected motion (vehicle-relative). Note that this is only possible because we can solve the absolute scale of both the vehicle and the drone poses using the LiDAR.
\section{Drone Detection and Tracking from LiDAR - Direct Observation} \label{sec:drone_tracking}
\subsection{LiDAR Scanning System}
Most LiDAR sensors \cite{vlp-16, hokuyo} perform rotational scanning along the vertical axis, i.e., varying azimuth angles, which makes them difficult to use for detecting flying objects such as drones. By reorienting the sensor such that the lasers rotate along a horizontal axis, i.e., varying elevation angles, we can overcome this limitation and detect objects that are located in the sky. To perform a complete spherical scan of the environment, we mount the LiDAR sensor on a motor that can rotate along the z-axis (see Fig. \ref{fig:overview}).

\begin{figure}[t]
    \centering
    \includegraphics[width=.6\linewidth]{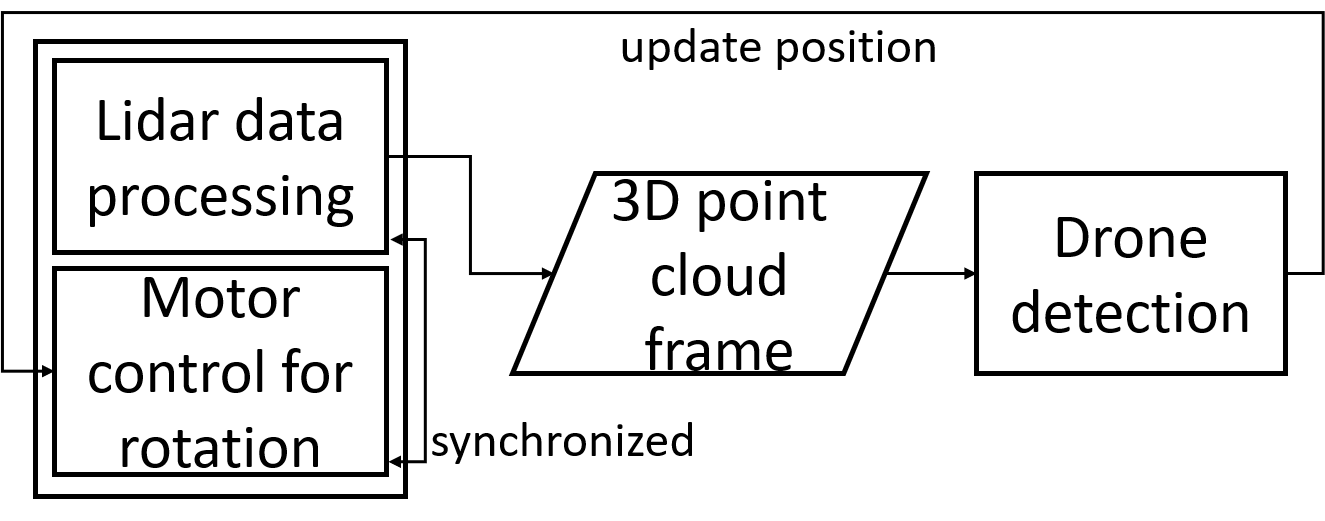}
    \caption{Tracking system process chart after initial scan. While rotating the LiDAR sensor, we capture 3D points from the environment aligned in one coordinate frame. We use the resulting 3D point cloud frame to run our proposed drone detection algorithm. We obtain the position of the detected drone, which we use to redirect the motor to this position and vibrate the motor around this position for further tracking of the drone.}
    \label{fig:process_chart}
\end{figure}

We generate a complete spherical scan of the scene by rotating the motor by at least $180^\circ$ and project the scan points to a sparse 2D depth image $I \in \mathbb{R}^{N \times N}$ of height $h$ parallel to the camera image plane. Multiple samples in a single pixel are filtered, and the smallest non-zero depth value is selected. 
Using this depth image, we perform an initial 2D detection of the drone using our dynamically adaptive kernel-based filtering. Then, we apply our 3D tracking algorithm to refine the detection and track the drone in real-time (see Fig. \ref{fig:process_chart}).

\subsection{Adaptive Kernel-based 3D Detector}
\label{subsec:adaptive}
\begin{figure}[t]
    \centering
    \includegraphics[width=.8\linewidth]{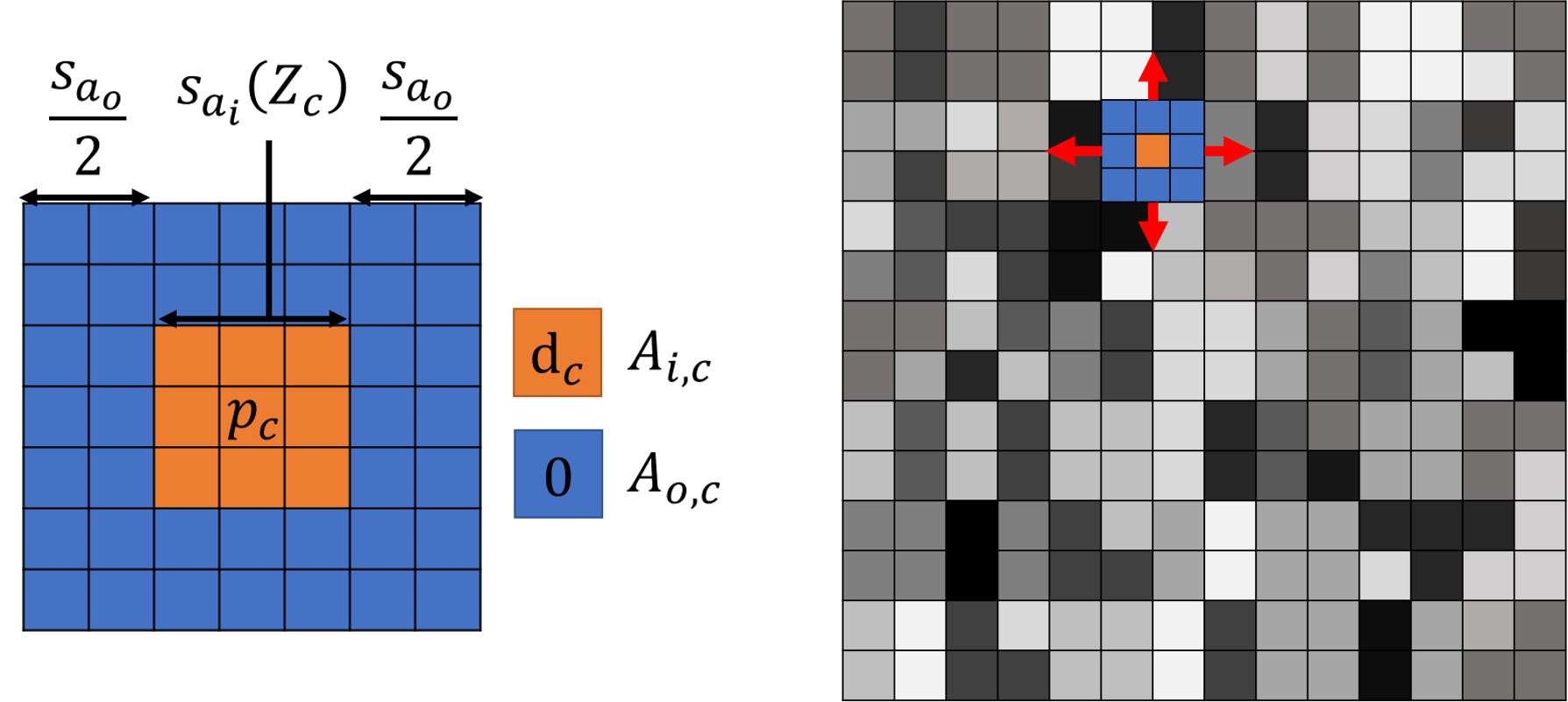}
    \caption{Adaptive kernel used for filtering (left) and the depth image's filtering process (right). The inner region of the kernel $A_{i,c}$ represents the drone and the outer region $A_{o,c}$ its immediate surrounding. By reasoning on the depth values within the kernel, we determine the region that resembles the drone and assign the center pixel of the region as the drone's center.}
    \label{fig:kernel}
\end{figure}

To detect the drone in the depth image, we define an adaptive kernel that dynamically changes in value and size depending on the expected height and dimension (in pixels) of the drone. 
The kernel consists of an inner and outer region, which represents the drone and its immediate surrounding, respectively (see Fig. \ref{fig:kernel}). 
The detection starts by assuming that each non-zero pixel in $I$ is a candidate for the drone's center $c$.
For each candidate pixel $p_c \in \mathbb{R}^2$, we calculate a dissimilarity measure $e(p_c)$ as the sum of the dissimilarity of the pixels in the inner part $e_i$ and outer part $e_o$ of the kernel:
\begin{equation} \label{eq:filter_value}
    e(p_c) = e_i(p_c) + e_o(p_c).
\end{equation}

We select the smallest dissimilarity value, which indicates the most probable position of the drone. The initial estimate of the GrD translation is the unprojection of the 2D hypothesis in 3D space and is solved by:
\begin{equation} \label{eq:min_equation}
    \hat{t}^0_{\GrD} = \text{unproject}(\argmin_{p \in \bar{I}} e(p))
\end{equation}
where $\bar{I}$ is the set of non-zero pixels of $I$.

\noindent
- \textbf{\emph{ Dissimilarity in Inner Region} }

The inner region is a non-zero detection kernel that indicates whether the pixels in the inner region $A_{i,c}$ have similar depth values to the depth $d_c$ of center pixel $p_c$. Formally, we define $e_i$ as:
\begin{equation} \label{eq:filter_value_i}
e_i(p_c)=\sum _{p\in A_{i,c}} |d(p)-d_c|. 
\end{equation}

The pixel size of the kernel's inner region within a depth image of resolution $res$ changes depending on the expected dimension of the drone of width $s$ at height $Z_c$. 
We calculate the dimension and set the size of the inner kernel as:
\begin{equation} \label{eq:inner_side_length}
    s_{a_i}(Z_c)=2 \cdot \left\lfloor\dfrac{s \cdot h/Z_c}{2 res}\right\rfloor +1.
\end{equation}
If $e_i$ is low, we can infer that there exists a cluster of points that is relatively the same size as the drone. 

\noindent
- \textbf{\emph{ Dissimilarity in Outer Region} }

The outer region assumes that the space around the inner region is empty when the drone is exactly in the inner region.
The dissimilarity $e_o(p_c)$ penalizes the pixels in the outer region $A_{o,c}$ that are similar to the assumed depth of the drone $d_c$. 
This means that when the inner region detects a drone, objects around it must either be behind or, in cases like occlusions, in front of the drone.

Accordingly, we define the outer region of the kernel as:
\begin{equation} \label{eq:filter_value_o}
e_o(p_c)= \sum _{p\in A_{o,c}}
\begin{cases}
0,&d(p)=0 \\
\dfrac{1}{\epsilon},&|d(p)-d_c|\leq \epsilon \\
\dfrac{1}{|d(p)-d_c|},&|d(p)-d_c| > \epsilon \\
\end{cases}
\end{equation}
where $\epsilon > 0$ is an arbitrary small number.
The outer region kernel further reinforces the inference of the inner kernel.

\subsection{Position Refinement and Real-time 3D Tracking}
We first refine the initial position estimate in Sec. \ref{subsec:adaptive} by calculating the center of the point cloud cluster of the drone within a spherical window of radius $r$ in 3D space. 
To do this, we adapt a simple iterative mean shift algorithm \cite{Yizong1995}.
Given the initial estimate's neighborhood (sphere with radius $r$) $F(\hat{t})=\left\{t : \left\| t-\hat{t}\right\| \leq r\right\}$ ($r>0$) and a 3D kernel function $C(t-\hat{t})=\exp(-\|t-\hat{t}\|^2)$,
the refined position is calculated per iteration $n$ as:
\begin{equation}
    \hat{t}^{n+1}_{\GrD} =\frac{\sum _{t \in F(\hat{t}^{0}_{\GrD})}C(t-\hat{t}^{n}_{\GrD}) \cdot t}{\sum _{t\in F(\hat{t}^{0}_{\GrD})}C(t-\hat{t}^{n}_{\GrD})}.
    \label{eq:meanshift}
\end{equation}
We run the mean shift algorithm for $n_{max}$ iterations and update $t_{\GrD} \leftarrow \hat{t}_{\GrD}$.

With the known initial position of the drone, we then proceed with tracking using the same mean shift algorithm. However, scanning the whole 3D space is slow and not suitable for real-time applications. 

Instead, to achieve real-time continuous tracking, we only slightly rotate the motor using a very small angle (vibration) around the expected direction of the drone to address the sparsity of the LiDAR's laser scan lines. We reorient the LiDAR to the currently detected direction and vibrate the motor between two angle values to generate a depth frame. For each frame, we perform a few mean shift iterations using (\ref{eq:meanshift}) to relocalize the drone and then reorient the center of the vibration for the next frame. We show a sample depth map of the vibration scanning in Fig. \ref{fig:drone_tracking}.
\section{Rotation Estimation using VPs - Indirect Observation} \label{sec:vp_detection}

Assuming that the environment has dominant parallel lines (e.g. buildings and roads), we can assume that the vehicle and the drone will detect vanishing points that are consensual to the general direction of these lines. 
By treating the detected VPs as vectors, i.e. absolute directions in world space, we can align these vanishing directions (VDs) and solve the relative rotation between the vehicle and drone.

In each camera coordinate system, we can define a 3x3 vanishing matrix containing three non-collinear VDs as column vectors i.e. $V = [v_1, v_2, v_3]$. We define these matrices $V_G \in \mathbb{R}^{3\times3}$ and $V_D \in \mathbb{R}^{3\times3}$ for the vehicle and the drone, respectively. Assuming that the matching VDs in $V_G$ and $V_D$ correspond to the same general direction in world space, we can calculate the ground vehicle - relative drone rotation $R_{G\rightarrow D}$, following \cite{Caprile1990}, as:
\begin{equation} \label{eq:rotation_drone_car}
    R_{G\rightarrow D} = V_G V_D^{-1}.
\end{equation}
Note that the VDs are simply the unprojection of the vanishing points, $\nu_i \in \mathbb{R}^2$, such that $v_{i} \propto K^{-1} [ \nu_{i}^T 1]^T$ where $K \in \mathbb{R}^{3 \times 3}$ is the intrinsic camera matrix. Generally, we can construct the vanishing matrix with only two VDs -- the third VD can be solved using the cross product: $v_3 = v_1 \times v_2$.

\noindent
- \textbf{\emph{Matching Vanishing Directions}}

Equation (\ref{eq:rotation_drone_car}) will give an accurate estimate of $R_{\GrD}$ only if the column vectors of $V_G$ and $V_D$ are corresponding VDs in world space. However, finding these correspondences is not straightforward, especially because the viewpoints are largely different.

In this method, we find the corresponding vanishing directions using the smallest angle approach. We first set an arbitrary initial $R_{G \rightarrow D0}$ (or use the value from a previous frame) and transform the individual VDs of the drone to the vehicle coordinate system. Then, we assign the correspondences as VDs with the smallest angle difference between them and rearrange the column vectors of $V_D$ to accommodate the changes. Finally, we calculate the relative rotation using Eq. (\ref{eq:rotation_drone_car}).

\noindent
- \textbf{\emph{Correcting Vanishing Directions Correspondences}}

Obviously, if $R_{G \rightarrow D0}$ is far from the actual value, the correspondences will be wrong. This problem becomes worse in a highly Manhattan world, where the three most dominant vanishing points are orthogonal. This means that if $R_{G \rightarrow D}$ has an error of greater than 45$^{\circ}$ around an axis, the aligned VPs will be matched to another axis and the smallest angle requirement will still be satisfied.

\begin{figure}
    \centering
    \includegraphics[width=.9\linewidth]{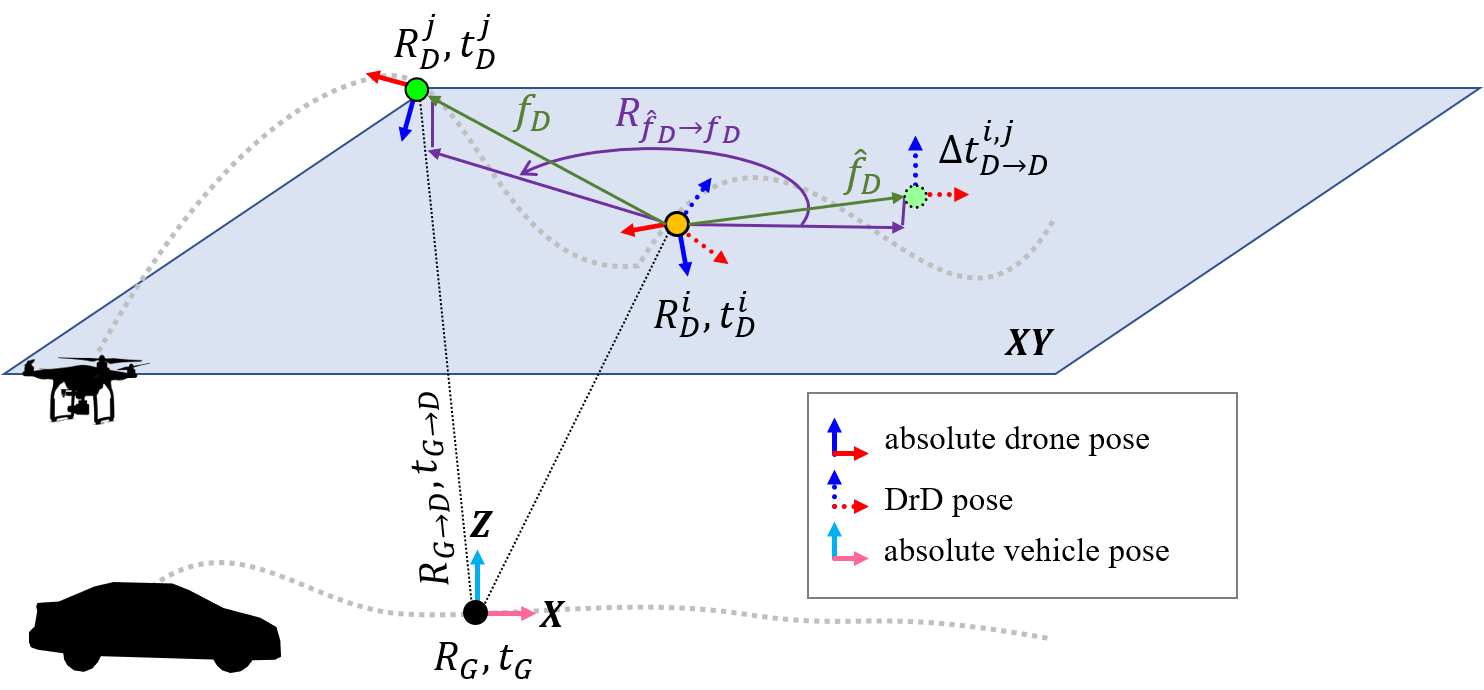}
    \caption{
    Translation vectors and rotation matrices in the different coordinate spaces. We correct the GrD rotation estimation by the rotation matrix $R_{\widehat{f}_D \rightarrow f_D}$, which aligns the $XY$-projected drone motion vectors as detected by itself $\widehat{f}_D$ and the vehicle $f_D$ in the world coordinate system.
    }
    \label{fig:translations_rotations}
\end{figure}

To address this problem, we reason on the motion of the drone as detected by itself and the vehicle in the world coordinate system, and find a correction (rotation) matrix $R_{\widehat{f}_D \rightarrow f_D}$.
We do this by aligning the motion vectors relative to its own coordinate system (drone-relative drone motion or DrD) to the one detected by the vehicle (see Fig. \ref{fig:translations_rotations}). Here, we assume that rotations around the horizontal axes are small enough and within the threshold of the smallest angle requirement. Therefore, we only need to explicitly correct the rotation around the vertical ($Z$) axis, as shown in Fig. \ref{fig:translations_rotations}.

Using two arbitrary frames $(i,j)$ with non-zero translation, the DrD translation $\Delta t_{D \rightarrow D}^{i,j}$ between $i$ and $j$ can be estimated using methods such as the 5-point algorithm \cite{Nister2003}.
Assuming the vehicle pose $R_G$ and $t_G$ of frame $i$ is known, the absolute drone translation vector, as observed by itself, is defined as:
\begin{equation}
    \widehat{f}_D = R_G \: R_{\GrD0} \: \Delta t_{D \rightarrow D}^{i,j}.
\end{equation}

We need to align vector $\widehat{f}_D$ to the motion vector $f_D$ of the drone as observed by the vehicle. Using the drone's translation vector $t_D = R_G \: t_{\GrD} + t_G$ in world space, we can simply solve this motion vector as:
\begin{equation} \label{eq:absolute_translation}
    f_D = t_D^j - t_D^i.
\end{equation}

We then obtain the correction matrix $R_{\widehat{f}_D \rightarrow f_D}$ by projecting $\widehat{f}_D$ and $f_D$ onto the $XY$ plane and solving the rotation matrix between the projected vectors.
Finally, the GrD rotation matrix can be corrected using:
\begin{equation} \label{eq:correct_rotation}
    R_{G \rightarrow D} \rightarrow R_G^{-1} R_{\widehat{f}_D \rightarrow f_D} R_G R_{G \rightarrow D0}.
\end{equation}
\section{Implementation}

\subsection{LiDAR-Motor Setup and Parameters}
\begin{figure}
   \centering
\begin{tabular}{cc}
\includegraphics[width=.4\linewidth]{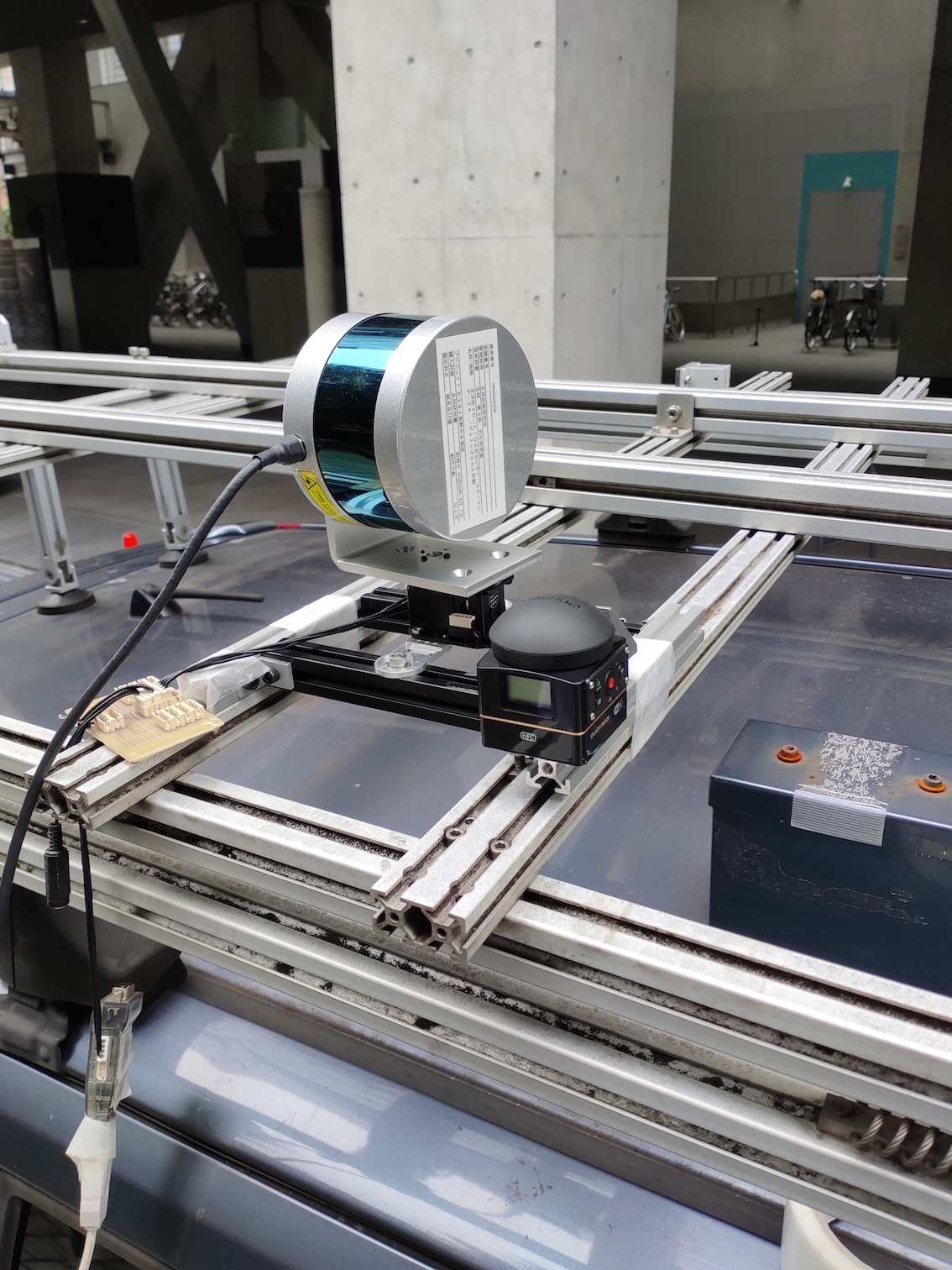}&
\includegraphics[width=.4\linewidth]{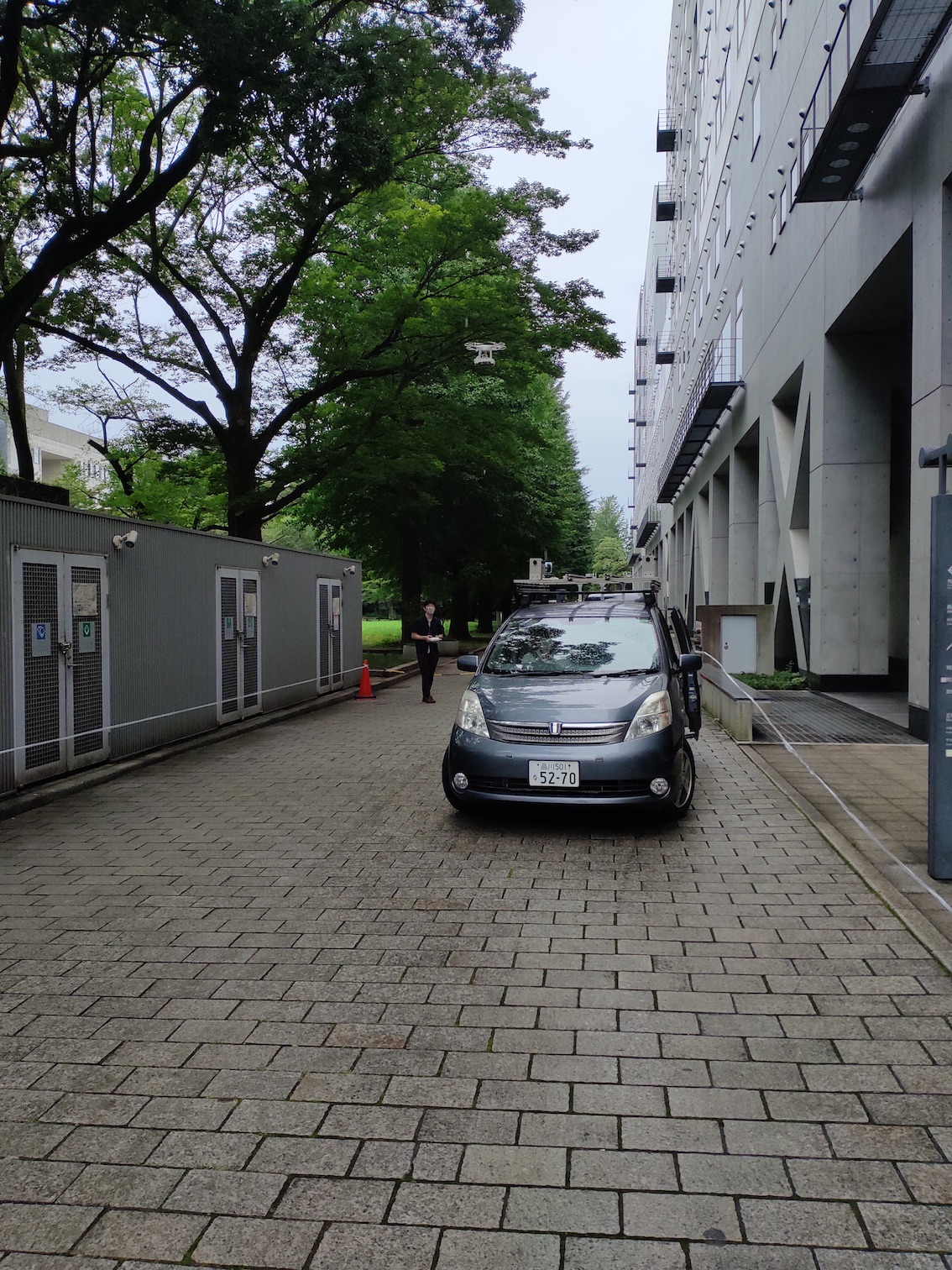}\\
\end{tabular}
\caption{Experimental setup showing LiDAR sensor and camera on vehicle's roof (left) and flying drone above vehicle (right)}
\label{fig:experimental_setup}
\end{figure}

The setup consists of a LiDAR sensor (Velodyne VLP-16 Puck (VLP-16) \cite{vlp-16}) connected to a servomotor (DYNAMIXEL MX-28AR \cite{dynamixel}) as shown in Fig. \ref{fig:experimental_setup}. 
The servomotor can rotate the LiDAR sensor by up to 360$^{\circ}$.
The whole upper hemisphere can be captured by rotating the servomotor by 180$^{\circ}$. We transform the measurement points by the LiDAR to compensate for the change in orientation accordingly.

For the projection parameters, we set the FOV $2 \theta_{max}=120^{\circ}$, $N=512$.
We set $s=0.5m$, $s_{a_o}=20$, and $\epsilon=0.1$ for the 2D detection kernel. 
For the mean shift algorithm, we set the number of iterations $n_{max}=10$ and the spherical neighborhood radius $r=1m$. We also set the servomotor's angular velocity to 11.4rpm for the initial 2D detection step and 57.2rpm for the vibration step with the amplitude of 5$^{\circ}$ and one frame generated per vibration period. Frame generation time can be increased or decreased depending on the speed of the motor, amplitude of the vibration, and LiDAR scanning speed. For our implementation, one frame is generated approx. every 120ms. We achieve a tracking computation time of 70ms per frame.

\subsection{LiDAR-Camera Fusion for Relative Ground Vehicle Pose Estimation}
First, feature points are extracted using Harris detector \cite{harris1988combined} and the extracted points are tracked among the frames with KLT tracker \cite{KLT_1981}. 
Next, we compute relative 5-DOF camera pose between $k$th and $k$th$+1$ camera frame using linear and non-linear processing. 

After calculating the 5-DOF parameters from the camera images, we determine the remaining scale parameter using pixels with depth values. We first establish 2D-3D correspondences by projecting LiDAR scan points onto the 2D images with the initial calibration parameters and track them to the other frames. Then we estimate the translation scale between $k$ and $k$th$+1$ frames using the 2D-3D correspondences. After that, the translation parameter is re-optimized by minimizing the re-projection error of the scanned points using bundle adjustment of the LiDAR points in the frame.

\subsection{Vanishing Point Detection}
We use one fisheye camera (KODAK PIXPRO SP360 4K VR Camera \cite{kodak}, FOV = 235$^{\circ}$) each for the drone and the car.
We apply a perspective decomposition on the images, choosing three distinct perspectives resulting in three rectified images and detect one VP in each view. In total, we obtain three non-collinear VPs per frame. For faster processing of the line detection, we resize the rectified images to 428x428. 

For VP detection, we first perform a line detection algorithm \cite{Lee2014}. Then, we divide the image into equally sampled square grids and choose the grid where most distinct lines pass through. We then run a single VP detection algorithm similar to \cite{Vanishing-Point-Detector} for the selected grid. 

The accuracy of the rotation estimation is mainly dependent on the accuracy of VP detection. To improve accuracy, we chose only to use the two best pairs of corresponding VPs. We also perform an extended Kalman filter for the GrD rotation to prevent large variation in rotations, which can result in the swapping of axes.

\subsection{Solving the Drone Trajectory}
We can obtain a reliable correction matrix $R_{\widehat{f}_D \rightarrow f_D}$ if we can accurately estimate the translation vectors $f_D$ and $\hat{f}_D$. However, these translation vectors are highly dependent on the accuracy of LiDAR detection and the relative pose estimation of the drone. The inaccuracy increases when the motion of the drone is very small or within the margin of error of the LiDAR scanning system. 

To address this problem, we sum a sequence of frames that have consistent motion direction (at maximum 30$^\circ$ angle difference). Taking the direction vectors of the last seven frames, we redefine (\ref{eq:absolute_translation}) as:
\begin{equation}
    f_D = \sum_{k = 0}^7 f_{D^k}.
\end{equation}
Additionally, to obtain more a reliable $f_{Dk}$, we choose distant frames to accumulate a longer translation. In this case, we set $j-i=14$ and a distance threshold of 1m.

It follows that if $f_D$ is reliable, we can assume that $\hat{f}_D$ is also reliable. In this case, we solve $\hat{f}_D$ in the same manner. To solve $\Delta t_{D \rightarrow D}^{i,j}$ for each drone frame, we use the 5-point algorithm \cite{Nister2003} with SURF as feature detector \cite{bay_tuytelaars_van}.
\section{Results and Discussion}

\begin{figure}
    \centering
    \begin{tabular}{cc}
        \includegraphics[width=.4\linewidth]{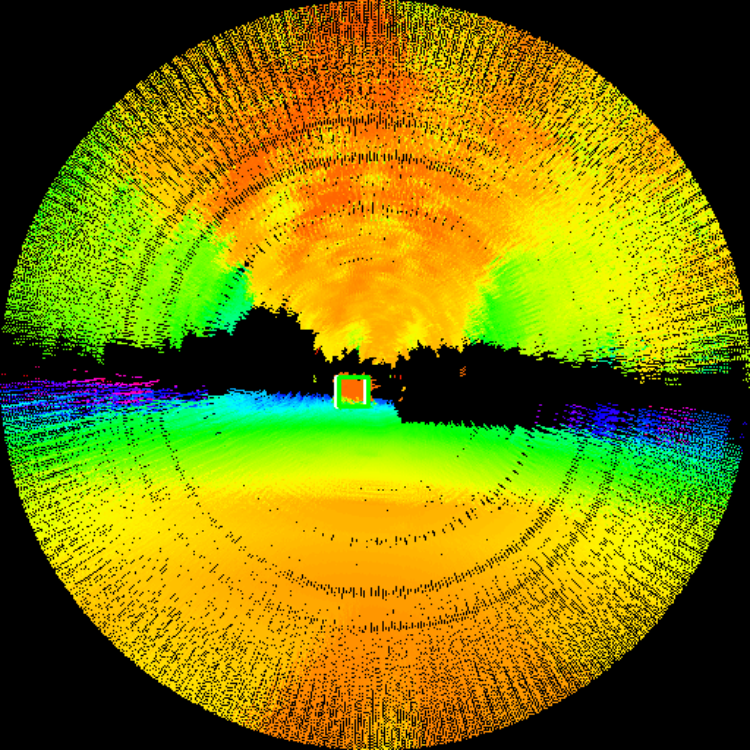}     &
        \includegraphics[width=.4\linewidth]{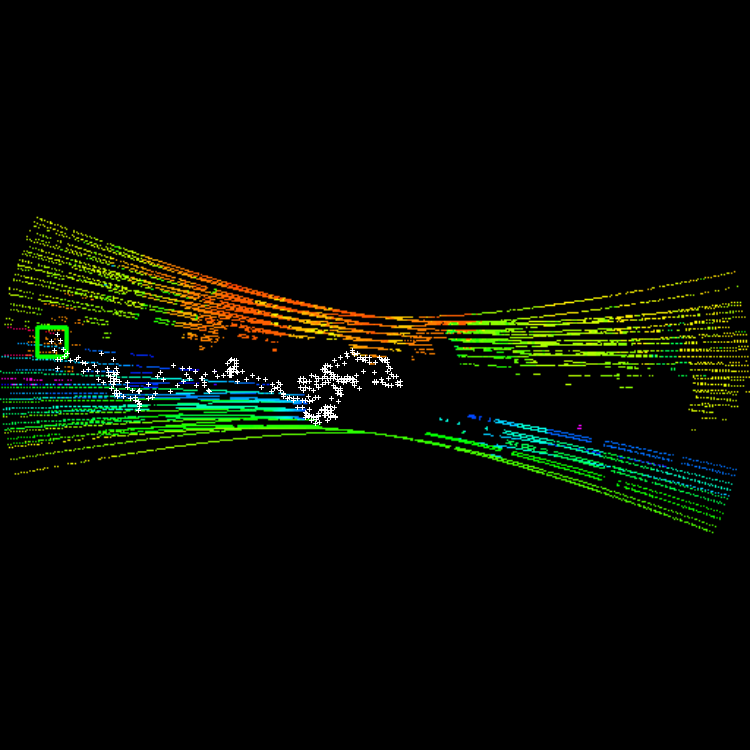}     \\
        Initial detection & Tracking trajectory
    \end{tabular}
    \caption{Depth of the scene with the detected drone (bounding box) and its tracked trajectory (white crosses). The left image shows the detection result of the initial scan. The right image shows the current depth image and detected drone at a certain point during tracking together with the tracked trajectory.}
    \label{fig:drone_tracking}
\end{figure}

\begin{figure*}[t]
    \centering
    \begin{tabular}{cccc}
        \includegraphics[width=.23\linewidth]{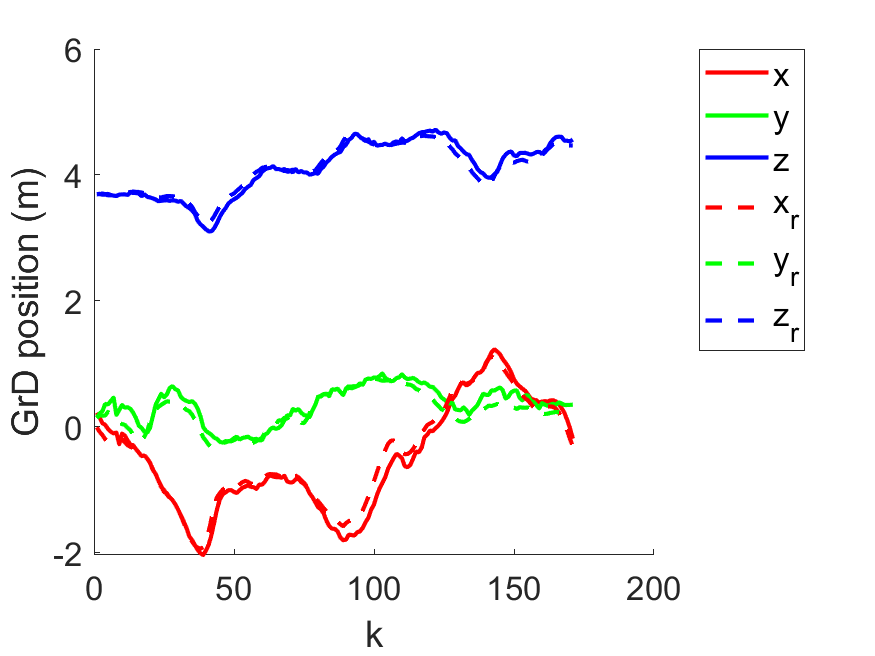}     &
        \includegraphics[width=.23\linewidth]{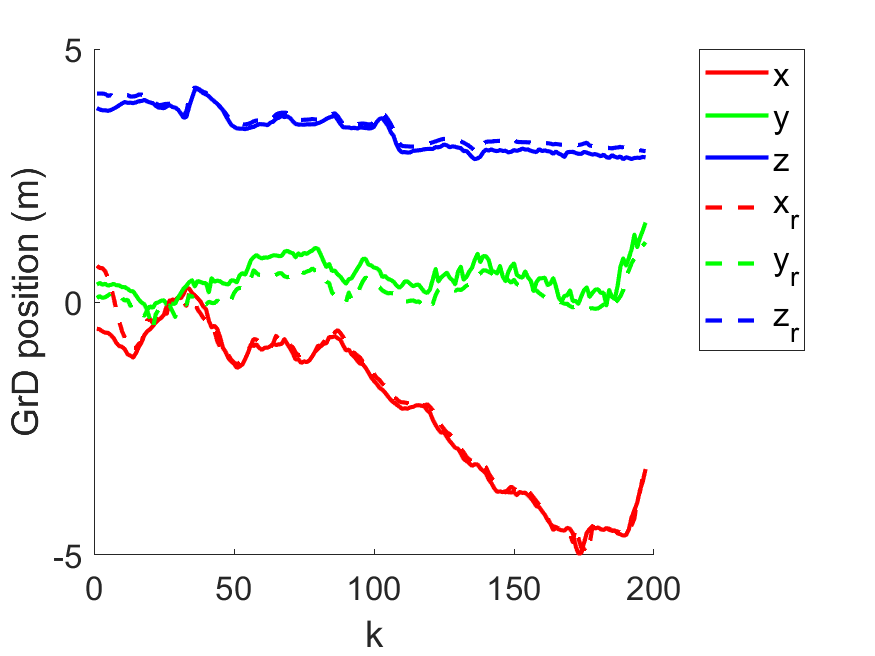}     &
        \includegraphics[width=.23\linewidth]{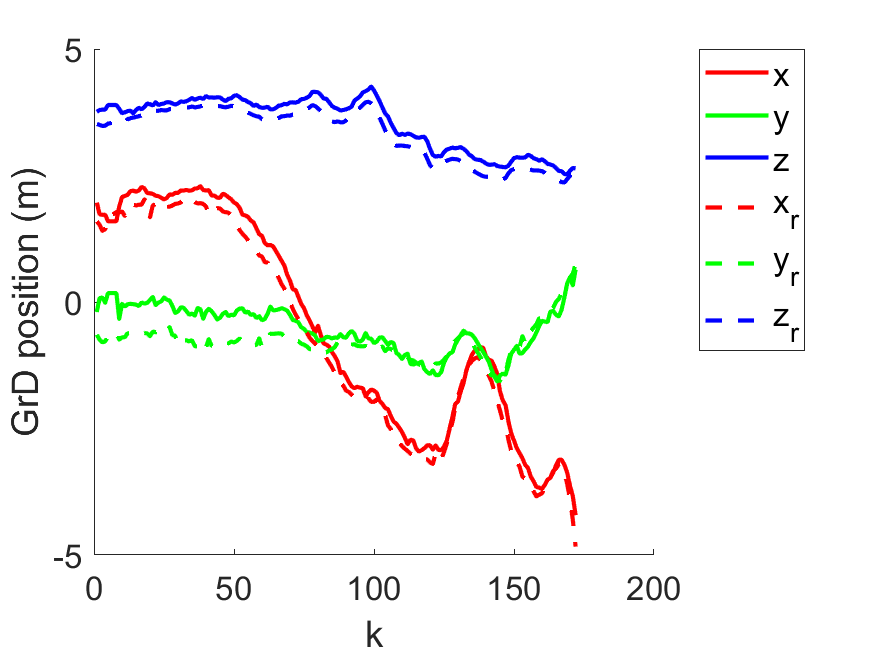}     &
        \includegraphics[width=.23\linewidth]{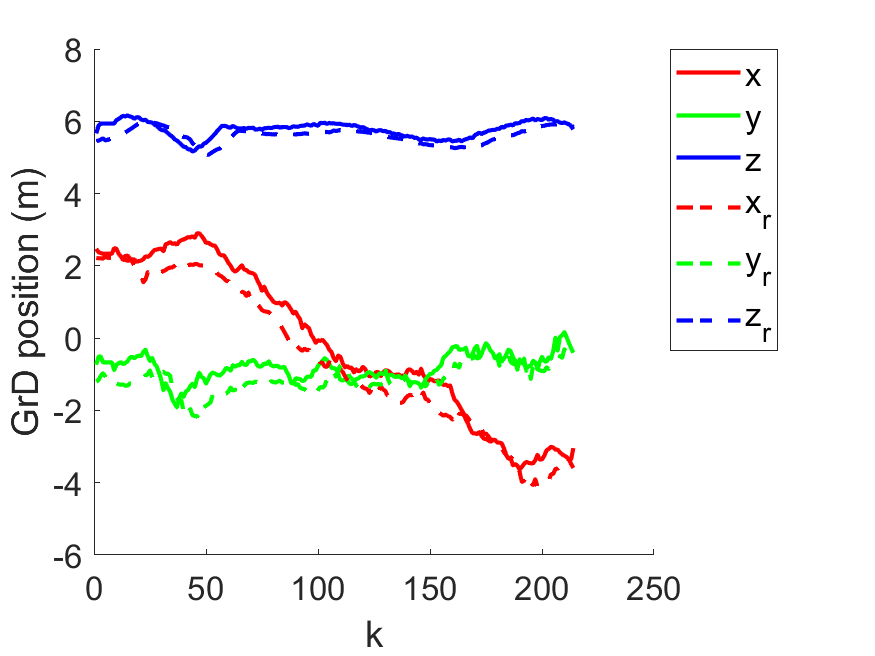}     \\
        \includegraphics[width=.23\linewidth]{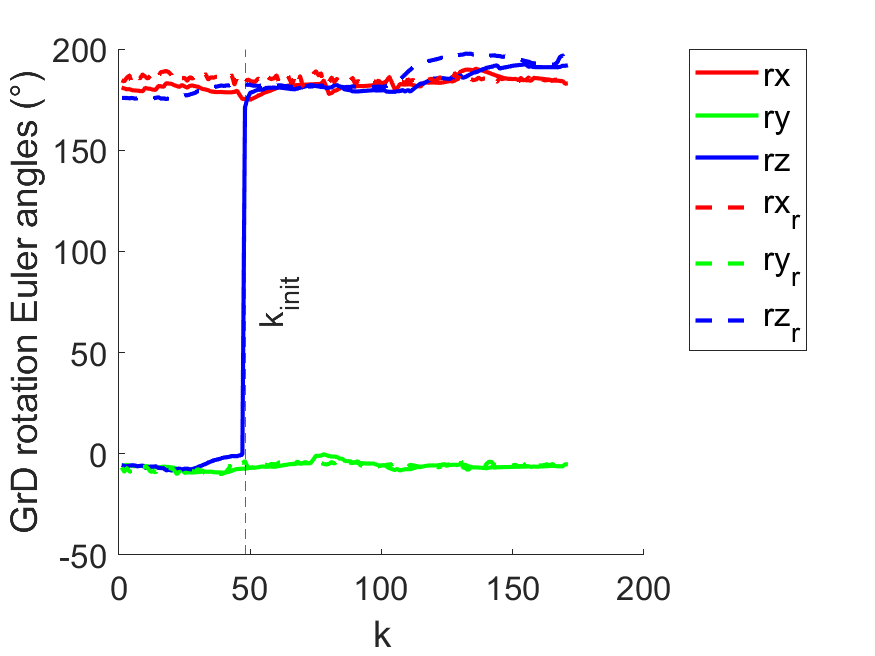}     &
        \includegraphics[width=.23\linewidth]{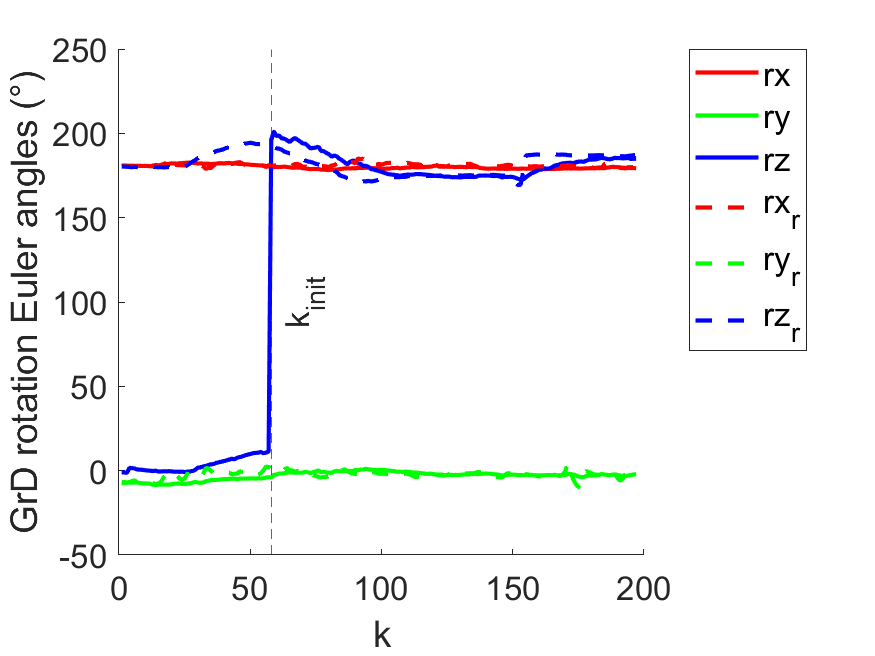}     &
        \includegraphics[width=.23\linewidth]{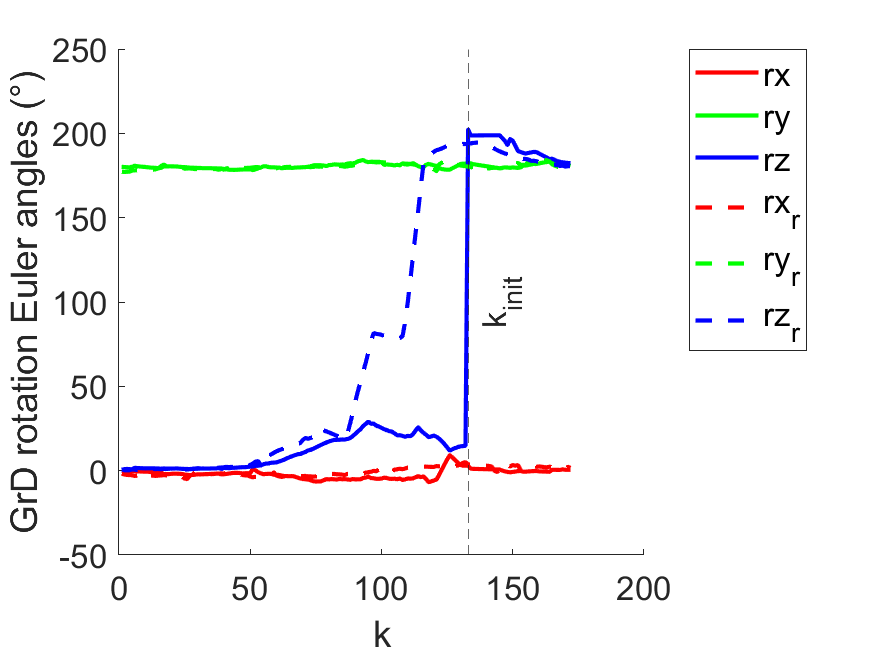}     &
        \includegraphics[width=.23\linewidth]{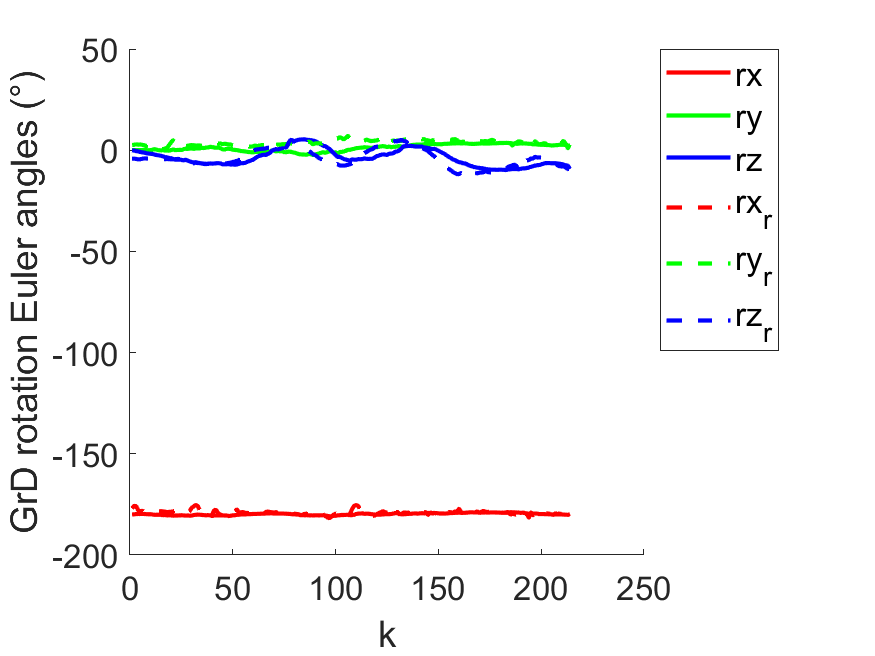}     \\
        Experiment 1 & Experiment 2 & Experiment 3 & Experiment 4
    \end{tabular}
    \caption{Experiment results: Comparison of GrD position (above) and rotation (below) with reference data (subscript $r$). When there is a big gap between the reference and estimated rotation, our proposed rotation correction algorithm (Sec. \ref{sec:vp_detection}) detects and corrects the rotation at $k_{init}$.}
    \label{fig:results}
\end{figure*}

\begin{table}
    \centering
    \begin{tabular}{|c|c|c|c|}
    \hline
    & $\Delta x$ & $\Delta y$ & $\Delta z$ \\
    \hline
    Exp. 1 & 0.1503m & 0.1511m & 0.0873m \\
    \hline
    Exp. 2 & 0.2498m & 0.3099m & 0.1372m \\
    \hline
    Exp. 3 & 0.2910m & 0.4117m & 0.2319m \\ 
    \hline
    Exp. 4 & 0.5519m & 0.4480m & 0.2403m \\  
    \hline
\end{tabular}
    \caption{Experiments' root-mean-square errors of position coordinates with respect to reference data}
    \label{tab:RMSD_pos}
\end{table}

We test our algorithm on real-world scenes using a car and a flying drone. In this paper, we present four experiments and test the accuracy of our proposed drone detection and tracking algorithm (Sec. \ref{sec:drone_tracking}) and orientation estimation using VPs (Sec. \ref{sec:vp_detection}). In each experiment, we allow the vehicle and drone to move independently from each other. The following evaluations are described in the vehicle's camera right-handed coordinate system, where the x-axis points towards the vehicle's front direction, y-axis to the left, and z-axis to the top (compare Fig. \ref{fig:translations_rotations}).

\subsection{Reference Data}
The reference data is obtained using a precise and colored dense 3D model obtained from laser range finder ZF Imager 5010C \cite{zf_imager} and aligned images from cameras on drone and vehicle. 
First, we calculate the camera motion without absolute scales using structure-from-motion implemented in MetaShape \cite{metashape}. 
Then, we manually select corresponding feature points in several fisheye images and the 3D model. Finally, the absolute scale and camera positions in the 3D model are optimized by minimizing the re-projection error of the 2D-3D correspondences.

\subsection{Accuracy of Drone Detection and Tracking}

A sample depth image for the initial drone detection, as well as the continuous tracking, is shown in Fig. \ref{fig:drone_tracking}. In this environment, objects that are sparse, such as trees, can be detected as the drone. Nevertheless, using our proposed method, we eliminated this problem by explicitly finding an object of the actual size and shape of the drone. From our results, we can see that even though sparse areas exist, and that the LiDAR points are very sparse, we can still successfully detect and continuously track the drone.

We show the results of our relative position estimation in Fig. \ref{fig:results}. We can see from the plots that we are able to achieve highly accurate GrD position estimates with an average error of less than 0.6m. We summarize the root-mean-square errors with respect to the reference values in Tab. \ref{tab:RMSD_pos}.

\begin{figure}
    \centering
    \begin{tabular}{cc}
        \includegraphics[width=.35\linewidth]{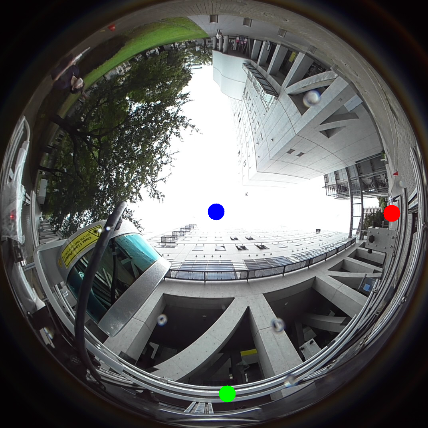} &
        \includegraphics[width=.35\linewidth]{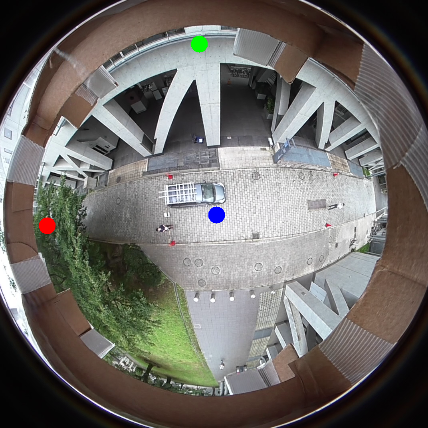}
   \end{tabular}
    \caption{Detected VPs in vehicle's (left) and drone's (right) fisheye lens images (same colors indicate corresponding VPs)}
   \label{fig:VPs}
\end{figure}

\begin{table}
    \centering
    \begin{tabular}{|c|c|c|c|}
    \hline
    & $\Delta rx$ & $\Delta ry$ & $\Delta rz$ \\
    \hline
    Exp. 1 & 3.6211$^{\circ}$ & 1.5556$^{\circ}$ & 5.5220$^{\circ}$ \\
    \hline
    Exp. 2 & 2.0632$^{\circ}$ & 1.7900$^{\circ}$ & 5.8581$^{\circ}$ \\
    \hline
    Exp. 3 & 2.3885$^{\circ}$ & 2.0870$^{\circ}$ & 5.2923$^{\circ}$ \\ 
    \hline
    Exp. 4 & 1.4057$^{\circ}$ & 2.9891$^{\circ}$ & 3.3560$^{\circ}$  \\  
    \hline
\end{tabular}
    \caption{Experiments' root-mean-square errors of Euler angles with respect to reference data after time step $k_{init}$ or, in case of experiment 4, $k = 0$}
    \label{tab:RMSD_rot}
\end{table}

\subsection{Accuracy of Orientation Estimation using VPs}

In the presented experiments, we show how our method can correct the alignment between VPs online. To set this up, we ran the VP alignment at the beginning of the experiments with an arbitrarily set rotation. Naturally, the VPs will align with the shortest angle difference. If the initial value is wrong, the rotation can lock on the wrong rotation (before $k_{init}$ in Fig. \ref{fig:results}).

After the movement of the drone is detected and a reliable estimate of the correction matrix is acquired, our algorithm updates the estimated rotation (after $k_{init}$ in Fig. \ref{fig:results}). From the results, we can see that that the estimated rotation is consistent with the reference data. The average error between estimation and reference data is less than 6$^{\circ}$. We summarize the rotation error in Tab. \ref{tab:RMSD_rot}.

In experiment 3, the VP rotation tracker failed to follow the fast rotating drone due to our Kalman filter implementation. We set the parameters of the KF to naturally clean up the highly inaccurate VP detection. Because of this, we set the KF to allow only a slow angular velocity. Nevertheless, our correction technique still detected the error and successfully corrected the rotation.

In experiment 4, the initial rotation is close to the correct rotation, and therefore, the correction matrix did not have to compensate for the estimation. We believe that the remaining inaccuracies in the rotation estimation are mainly due to inaccurate VP detection.
\section{CONCLUSIONS}
We presented a cost-effective ground vehicle - drone relative pose estimation demonstration system using a LiDAR sensor mounted on a rotating motor and two fisheye cameras. The system is fully-automatic, e.g., it can recover the pose in the middle of deployment by reasoning on the relative motions between the two cameras. To the best of our knowledge, this is the first relative pose estimation method with scale between a drone and a vehicle utilizing a single LiDAR sensor and two cameras. 

Our experiments showed that we can successfully detect and track the relative pose between a ground vehicle and a drone. However, for future work, there are several improvements that can be done. The drone detection algorithm can be extended by considering occlusion handling or distinguishing drones from other similar objects. To improve the accuracy of the rotation estimation, a more reliable VP detection algorithm can be used, for example, by detecting multiple VPs directly in a fisheye lens image. Additional sensors can also be added, such as gravity sensors and IMUs.






\bibliographystyle{IEEEtranBST/IEEEtran}
\bibliography{IEEEtranBST/IEEEabrv,mybibfile}

\end{document}